\documentclass[review,12pt,sort&compress]{elsarticle}

\usepackage{epsfig}
\graphicspath{ {./Figures/} }
\usepackage[rightcaption]{sidecap}
\usepackage{wrapfig}
\usepackage{adjustbox}
\usepackage{multirow}
\usepackage{lscape}
\usepackage{graphicx}
\usepackage[pass]{geometry}
\usepackage{vcell}
\usepackage{url}
\usepackage{amssymb}

\usepackage{lineno}

\journal{XXXXX}

\begin{document}

\begin{frontmatter}



\title{Ethics of Artificial Intelligence and Robotics in the Architecture, Engineering, and Construction Industry}


\author[inst1]{Ci-Jyun Liang}
\author[inst2]{Thai-Hoa Le}
\author[inst3]{Youngjib Ham}
\author[inst4]{Bharadwaj R. K. Mantha}
\author[inst5]{Marvin H. Cheng}
\author[inst2]{Jacob J. Lin*}

\affiliation[inst1]{organization={Department of Civil Engineering, Stony Brook University},
            addressline={2434 Computer Science},
            city={Stony Brook},
            state={NY},
            postcode={11794},
            country={USA}}

\affiliation[inst2]{organization={Department of Civil Engineering, National Taiwan University},
            addressline={No.1, Sec. 4, Roosevelt Road}, 
            city={Taipei},
            postcode={10616}, 
            country={Taiwan}}

\affiliation[inst3]{organization={Department of Construction Science, Texas A\&M University},
            addressline={574 Ross Street},
            city={College Station},
            state={TX},
            postcode={77843},
            country={USA}}
            
\affiliation[inst4]{organization={Department of Civil and Environmental Engineering, University of Sharjah},
            addressline={P.O. Box 27272},
            city={Sharjah},
            country={UAE}}

\affiliation[inst5]{organization={School of Engineering, Embry-Riddle Aeronautical University},
            addressline={1 Aerospace Boulevard},
            city={Daytona Beach},
            state={FL},
            postcode={32114},
            country={USA}}

\begin{abstract}
Artificial intelligence (AI) and robotics research and implementation emerged in the architecture, engineering, and construction (AEC) industry to positively impact project efficiency and effectiveness concerns such as safety, productivity, and quality.
This shift, however, warrants the need for ethical considerations of AI and robotics adoption due to its potential negative impacts on aspects such as job security, safety, and privacy. Nevertheless, this did not receive sufficient attention, particularly within the academic community. This research systematically reviews AI and robotics research through the lens of ethics in the AEC community for the past five years.
It identifies nine key ethical issues namely job loss, data privacy, data security, data transparency, decision-making conflict, acceptance and trust, reliability and safety, fear of surveillance, and liability, by summarizing existing literature and filtering it further based on its AEC relevance.
Furthermore, thirteen research topics along the process were identified based on existing AEC studies that had direct relevance to the theme of ethics in general and their parallels are further discussed.
Finally, the current challenges and knowledge gaps are discussed and seven specific future research directions are recommended. This study not only signifies more stakeholder awareness of this important topic but also provides imminent steps towards safer and more efficient realization.

\end{abstract}




\end{frontmatter}


\section{Introduction}
\label{sec:intro}


The Architecture, Engineering, and Construction (AEC) industry has been continually growing over the last decade, which is one of the most influential markets globally \citep{timetric_global_2016}.
Employment in the U.S. construction industry has been growing between 2011 and 2019 \citep{brown_fatal_2021}.
Even during the COVID-19 pandemic unemployment rate surge, the decreasing rate of construction industry employment is lower than the overall industry employment rate, and is almost bounced back to pre-pandemic \citep{harris_employment_2022}.
Despite its growth, the AEC industry still has critical issues related to safety, productivity, and quality \citep{liang_human-robot_2021}.
According to the Center for Construction Research and Training (CPWR) and the U.S. Bureau of Labor Statistics (BLS) report, the number of fatal and nonfatal injuries in the construction industry has been increasing from 2011 to 2020 \citep{trueblood_fatal_2022}.
In addition, productivity and quality still remain the significant long-term challenges in the AEC industry \citep{shehzad_building_2019,lundeen_autonomous_2019}.

Researchers in the AEC industry have explored emerging technologies such as Artificial Intelligence (AI) and robotics to improve safety, productivity, and quality control throughout the project life-cycle.
During the design phase, AI technologies have shown promise on issues such as clash prediction \citep{hu_clash_2019} or prefabrication \citep{baghdadi_design_2020}, and cost prediction \citep{cao_prediction_2018,garcia_de_soto_productivity_2018}.
During the construction phase, studies showed significant positive implications on safety and productivity.
For example, computer vision methods were initially used to locate human workers \citep{jeelani_real-time_2021} and heavy equipment on jobsites \citep{liang_vision-based_2019}.
Then, these were advanced further to identify hazards \citep{zhang_recognition_2020} and also analyze productivity \citep{chen_automated_2020}.
Scene understanding and reconstruction methods were observed to be critical for construction progress monitoring \citep{lundeen_scene_2017,lin_construction_2020}.
Robotized equipment is also deployed on construction sites to mitigate heavy-duty tasks like bricklaying, assembly, or demolition \citep{ding_bim-based_2020,liang_ras:_2017,yang_collision_2021,mu_intelligent_2022}.
Similarly, during the operation and maintenance phase, AI and robotics showed promise to effectively inspect and monitor the built environment \citep{mo_automated_2020,zhao_structural_2021,xu_occupancy_2019,mantha_robotic_2018}.

With growing attention to AI and robotics in AEC research, systematic and state-of-the-art review methods have been applied to analyze existing literature.
\citet{darko_artificial_2020} utilized scientometric analysis to examine research trends in AI in the AEC industry and provided future research recommendations.
\citet{chen_construction_2018} reviewed related research in construction automation using the text mining method.
\citet{emaminejad_trustworthy_2022} conducted a systematic review of AEC literature to investigate trustworthy AI and robotics issues.
\citet{liang_humanrobot_2021} proposed a taxonomy to categorize literature on human-robot collaboration in construction and suggested future research directions.
\citet{ham_visual_2016} surveyed relevant AEC research in aerial robots and visual monitoring and discussed the potential of automatic monitoring and inspection.
However, none of the above-mentioned studies include ethical concerns.

Ethical concerns in AI and robotics have been accentuated recently \citep{torresen_review_2018,dubber_oxford_2020,bartneck_introduction_2021}.
Ethics is an important aspect that affects the workplace and society \citep{van_wynsberghe_ethical_2022,dubber_perspectives_2020}, which has also been taken into account in the AEC industry.
Researchers have been debating the ethical issues and dilemmas in architecture and construction \citep{adnan_ethical_2012,ray_architecture_2005,bowen_ethical_2007,li_impact_2022,oladinrin_interrelations_2023}.
However, such ethical concerns in the AEC industry do not consider the presence of AI and robotics.
For example, the rapidly growing large language model (LLM), e.g., ChatGPT \citep{radford_language_2019}, has changed methods to complete job duties in some industries.
Researchers have started to examine the use of ChatGPT in AEC, such as construction management \citep{prieto_investigating_2023}.
This raises concerns about job replacement or security issues when applying these emerging technologies.
Therefore, it is necessary to investigate the ethical issues of AI and robotics, particularly in the context of the AEC industry.

The objective of this research is to conduct a systematic review of AI and robotics research in the AEC discipline and select the relevant literature to analyze its ethical-related implications.
The general ethics of AI and robotics are reviewed first and used to define the ethical issues in AEC.
Next, the relevant literature is categorized into each ethical issue and discusses possible solutions.
Finally, future research directions in the ethics of AI and robotics in AEC are suggested.

The remainder of this article is arranged as follows.
First, the research methodology of this review is introduced.
Second, the general ethics of AI and robotics are reviewed and used to define the ethical issues of AI and robotics in AEC.
Third, keywords definition, source of the literature, and inclusion criteria are determined.
Fourth, the detailed ethical issues are discussed based on the systematic review results.
Lastly, future research directions are recommended according to the ethical issues.

\section{Research Methodology}
\label{sec:method}

\begin{figure}[htbp]
\centering
\includegraphics[scale=0.5]{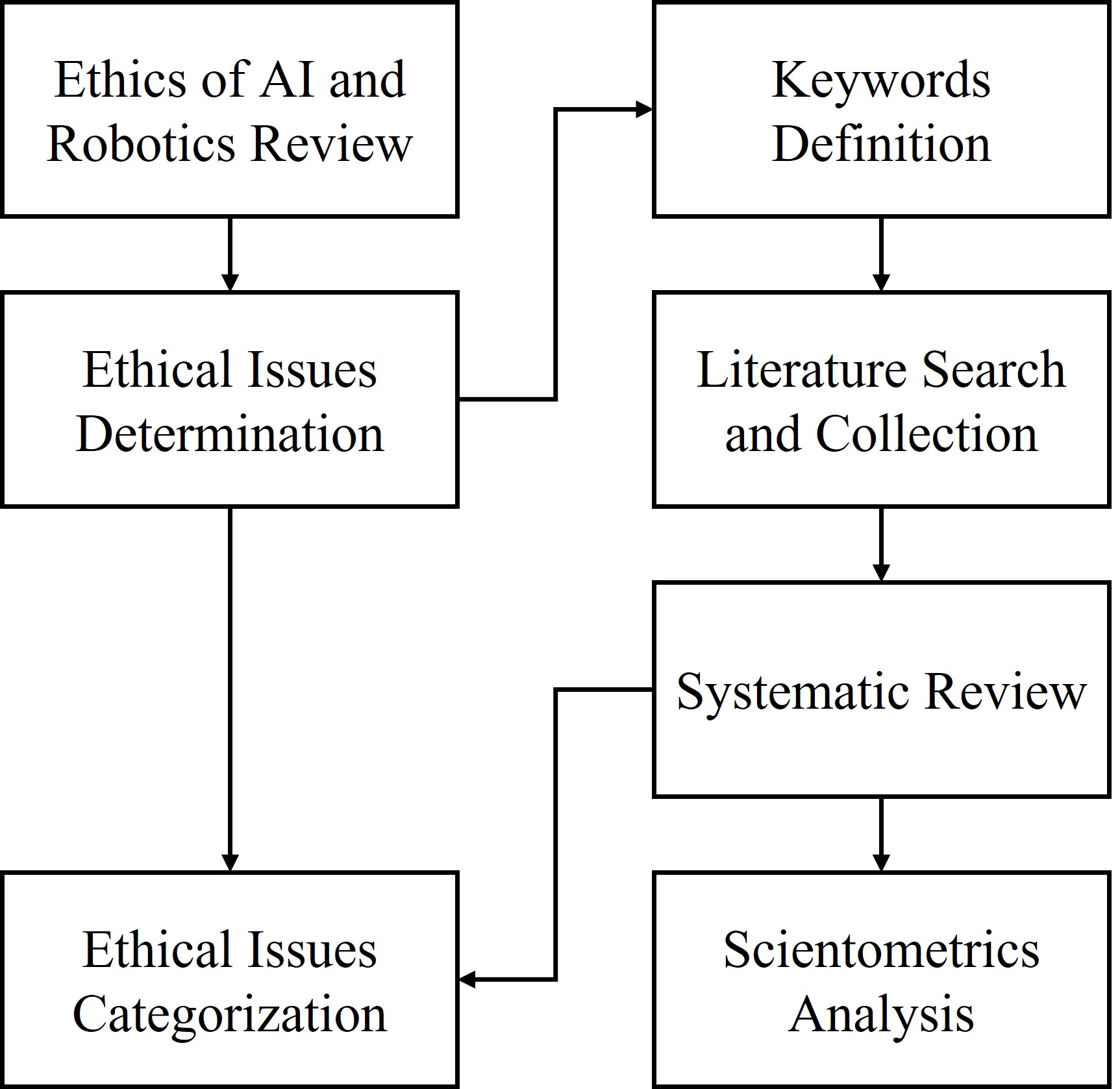}
\caption{Research methodology and steps}
\label{fig:Method}
\end{figure}

In this research, we conduct a systematic review to study the ethical issues of AI and robotics in the AEC discipline.
Figure~\ref{fig:Method} illustrates the research methodology and steps.
Systematic reviews are essential to effectively study a contemporary scientific field.
The method requires authors to be accurate, transparent, and up-to-date in reviewing completed research \citep{moher_reporting_2018}.
The reviewing results can provide scientists with reliable information on current research.
In order to optimize the research activities, PRISMA (Preferred Reporting Items for Systematic Review and Meta-analysis) is preferred.
The PRISMA statement, first published in 2009 and updated in 2020 (hereafter referred to PRISMA statement 2020 as PRISMA\footnote{http:/www.prisma-statement.org}), includes a 27-item checklist, explanation and elaboration, and a flow diagram intended to help researchers report comprehensive systematic reviews and meta-analyses  \citep{moher_reprint_2009,page_prisma_statement_2021}.
Those 27 recommended items in the checklist focus on the title, abstract, introduction, methods, results, discussion, and other information, such as supportive resources, competing interests, registration, protocol, etc.
An explanation and elaboration document is encouraged for those unfamiliar with PRISMA as it explains in detail the reasons for using suggested items \citep{page_prisma_explanation_2021}.
The flow diagram contains the identification, screening, and other included data that are related to the research scope. 

Since having advantages in systematic research, numerous scholars have employed this particular methodology across diverse domains to analyze substantial volumes of information.
For instance, \citet{zorzela_prisma_2016} developed a PRISMA harms checklist for health intervention reports based on the PRISMA statement, PRISMA for abstracts, and PRISMA for protocols.
Meanwhile, \citet{page_reproducible_2018} utilized PRISMA for meta-analyses to evaluate the frequency of reproducible research in biomedical studies.
\citet{leclercq_meta_2019} recommended PRIMA as an effective tool for reporting meta-analyses, with 87\% of the 206 meta-analyses explicitly following PRISMA.

\section{Ethics of AI and Robotics}
\label{sec:ethicsAIR}
Ethics is a set of moral principles and norms that govern an individual or a group of people to achieve good outcomes, according to Merriam-Webster dictionary\footnote{https://www.merriam-webster.com/dictionary/ethic}.
Ethics provides rules of conduct to society and encourages members to behave in a way that is right, rather than enforced by regulations like government laws.
Specifically, ethics examines human behaviors in terms of good and bad, or morally correct and wrong \citep{bartneck_introduction_2021}.
The ethical principles are determined and developed by society over the years \citep{shafer-landau_ethical_2012}, and all members of the society follow ethical behavior to build trust with each other.

As AI and robotics technologies are introduced to society, ethical concerns are also debated.
\citet{dubber_perspectives_2020} indicated that AI and robots are considered members of society since they are capable of making decisions, and we should investigate their impact on humans.
One of the famous ethical dilemmas, the trolley problem \citep{thomson_trolley_1985}, is incorporated into AI and robotics ethical discussions, which also leads to alternate challenging scenarios, e.g., The Molly Problem \citep{ITU_self_2020}.
Human-robot relationships are typically built upon human-human relationships because of their intelligent abilities \citep{de_graaf_ethical_2016}.
For example, human-to-human social cues and signals were shown in the human-robot interaction \citep{fiore_toward_2013}.
How to measure the impact of AI and robots on human society and how to minimize the negative impacts are crucial topics \citep{torresen_review_2018}.

Addressing the ethical challenges of AI and robotics has been studied in the past.
The most well-known three (four) laws of robotics formulated by \citet{asimov_three_1942} in 1942 and 1983 became the foundation of later AI and robotics ethical research.
Roboethics is the term defined in the Springer Handbook of Robotics to discuss the ethical issues of robotics \citep{siciliano_springer_2008,veruggio_roboethics_2016}.
A taxonomy of robotethics was proposed in the chapter to identify ethical issues of different robot applications.
\citet{torresen_review_2018} argued that the ethical issues of AI and robotics should be tackled in two ways, i.e., developing systems with ethical challenges-awareness and ethical decision-making ability.
An AI system with morality-based decision-making capability is an ideal solution, but it is extremely challenging to create such a system due to the existence of ethical dilemmas.
On the other hand, technology companies and councils have begun to investigate the ethical issues of AI and robotics.
For example, geospatial data are collected widely by the industry and used for AI applications.
This situation has been pointed out by the World Geospatial Industry Council and encouraged the discussion of the ethical use of such geospatial data \citep{wgic_geospatial_2020}.
Additionally, companies such as IBM have conducted AI ethics research to address the concerns of AI and personal data usage \citep{berger_addressing_2022,berger_AI_2023}.

Governments and standard committees are developing laws and standards to regulate the use of AI and robotics.
\citet{hacker_explainable_2020}\citep{hacker_varieties_2022} discussed AI ethics from the perspective of law.
New laws and regulations are created and necessitated by emerging AI cases.
The need for sustainable AI regulation and law has drawn legal communities' attention.
European Union has proposed the AI Act law to regulate AI applications \citep{eu_ai_2021}.
Those applications with unacceptable risks are prohibited, and applications with high risks are regulated.
Similarly, the European Union's Digital Services Act (DSA) and General Data Protection Regulation (GDPR) govern the privacy and security of personal data used online or in AI systems.
Various associations and committees are discussing and proposing new standards for ethical AI and robots.
The British Standards Institute proposed a standard for the ethical design of robots \citep{bsi_robots_2016}.
Ethical risks should be assessed and measured when designing robots, which is similar to the safety assurance in robot design.
The Institute of Electrical and Electronics Engineers (IEEE) Standards Association launched an initiative on the ethics of autonomous and intelligent systems \citep{ieee_ethically_2019}.
Several IEEE standard committees are working to address the ethical issues in technology, i.e., IEEE P7000\texttrademark series.

To identify the ethical issues of AI and robotics in the AEC industry, we first summarize the general ethical issues of AI and robotics discussed in previous literature.
\citet{lin_robot_2011} identified three categories of robot ethical issues, which are safety and errors, law and ethics, and social impact.
First, safety and errors include robot failures and cybersecurity concerns.
Second, law and ethics are related to liability, responsibility, and privacy.
Programming robots to follow the law and ethics code is one solution, but it is still unclear how to achieve it in practice.
Third, social impact includes job loss, skill loss, and emotional impact.
\citet{van_wynsberghe_ethical_2022} categorized ethical issues of industrial robots into eight categories: "job loss and reorganization of labor," "informed consent, data collection, and privacy," "user-involved design," "hierarchical decision-making," "acceptance and trust," "psychological harm," "emotional impact," and "performance monitoring (fear of surveillance)."
User-involved design refers to involving end-users during the design of robotic systems so that they can understand how the systems work and build trust relationships.
Hierarchical decision-making refers to the authority in human-robot collaboration and who is responsible for making the final decision.
Performance monitoring refers to surveillance concerns in the workplace, also known as the chilling effect.
Workers might feel they are being watched by collaborative robots and being evaluated on their work performance.

\citet{veruggio_roboethics_2016} introduced the concept of roboethics and discussed ethical issues of different types of robots in a taxonomy.
Job loss, privacy, morality, rights and responsibilities, emotional relationships, instruction conflicts, cybersecurity, unpredictable behaviors, liability, and psychological problems were discussed among various robot applications.
Five recommendations were made during the Euron Roboethics Atelier 2006 \citep{veruggio_euron_2006}: safety, security, traceability, identifiability, and privacy.
\citet{dubber_oxford_2020} edited the Oxford Handbook of Ethics of AI, which debated the ethical issues of AI from different perspectives.
The handbook recognized fairness, accountability, transparency, responsibility, labor displacement, rights and well-being, autonomy, and sexuality as general issues related to AI ethics.
Similarly, the introductory book written by \citet{bartneck_introduction_2021} introduced five topics of AI and robotics ethics, which are trust and fairness, responsibility and liability, risk in business, psychological aspects, and privacy.
\citet{berger_AI_2023} discussed the current AI ethics issues, including data privacy and governance, explainability and trust, accountability, fairness, profiling and manipulation, and social impact.

Trustworthy AI is a concept to build the trust of AI in individuals and societies using ethical concepts.
Five principles were introduced by \citet{thiebes_trustworthy_2021}: beneficence, non-maleficence, autonomy, justice, and explicability.
\citet{liu_trustworthy_2022} also proposed six dimensions of trustworthy AI, which are safety and robustness, nondiscrimination and fairness, explainability, privacy, accountability and auditability, and environmental well-being.
\citet{emaminejad_trustworthy_2022} reviewed trustworthy AI and robotics research in the AEC industry.
Explainability and interpretability, performance and robustness, reliability and safety, and privacy and security are four identified categories of trust dimensions.
Finally, the European Commission has published ethics guidelines for trustworthy AI, defining four ethical principles and seven requirements \citep{ai_hleg_ethics_2019}.
The four ethical principles are respect for human autonomy, prevention of harm, fairness, and explicability.
The seven requirements include human agency and oversight, technical robustness and safety, privacy and data governance, transparency, diversity and fairness, societal and environmental well-being, and accountability.
These requirements are later used to develop an assessment list for trustworthy AI (ALTAI) \citep{ai_hleg_assessment_2020}.
The detailed items of the ALTAI are listed in Table \ref{table:altai}.

\begin{table}
\caption {Assessment list for trustworthy AI \citep{ai_hleg_assessment_2020}}
\label{table:altai}
\begin{tabular}{ll}
\hline
Requirement & Item \\ \hline
\multirow[t]{2}{*}{Human agency and oversight} & Human agency and autonomy \\ & Human oversight \\
\multirow[t]{4}{12em}{Technical robustness and safety} & Resilience to attach and security \\ & General safety \\ & Accuracy \\ & \begin{tabular}[c]{@{}l@{}}Reliability, fall-back plans, and\\ reproducibility\end{tabular} \\
\multirow[t]{2}{*}{Privacy and data governance} & Privacy \\ & Data governance \\
\multirow[t]{3}{*}{Transparency} & Traceability \\ & Explainability \\ & Communication \\
\multirow[t]{3}{*}{Diversity and fairness} & Avoidance of unfair bias \\ & Accessibility and universal design \\ & Stakeholder participation \\
\multirow[t]{3}{15em}{Societal and environmental well-being} & Environmental well-being \\ & Impact on work and skills \\ & \begin{tabular}[c]{@{}l@{}}Impact on society at large or\\ democracy\end{tabular} \\
\multirow[t]{2}{*}{Accountability} & Auditability \\ & Risk management \\ \hline
\end{tabular}
\end{table}

Based on the general ethical issues of AI and robotics, we define the ethical issues of AI and robotics in the AEC discipline.
We begin by comparing the above-mentioned ethical issues and merging similar items.
We then determine whether the specific issue is related to the AEC industry scenario.
As a result, we recognize nine categories of ethical issues of AI and robotics in AEC, which are "Job Loss," "Data Privacy," "Data Security," "Data Transparency," "Decision-Making Conflict," "Acceptance and Trust," "Reliability and Safety," "Fear of Surveillance," and "Liability."
Table \ref{table:ethics} shows the nine AEC AI and robotics ethical issues.
Note that we do not consider "sexuality," "diversity," and "fairness, justice, beneficence, and nondiscrimination" as ethical issues of AI and robotics in AEC in our research since they are not related to the AEC industry.


\begin{landscape}
\begin{table}
\centering
\caption{Ethical issues from literature}
\label{table:ethics}
\resizebox{\paperwidth}{!}{%
\begin{tabular}{llllllllll} [!hbt]\\
\hline
Reference  & \multicolumn{9}{c}{Ethical issues}\\ 
\hline
Lin et al. [61]  & Job and skill loss  & Privacy  & Cybersecurity  &   & Emotional impact  & Robot failures  &  &  & \begin{tabular}[t]{@{}l@{}}Liability and\\responsibility\end{tabular} \\ 
\hline
\begin{tabular}[t]{@{}l@{}}van Wynsberghe\\et al. [33]\end{tabular} & \begin{tabular}[t]{@{}l@{}}Job loss and\\reorganization\\of labor\end{tabular} & & \begin{tabular}[t]{@{}l@{}}Informed consent, data\\collection, and privacy\end{tabular} & User-involved design & \begin{tabular}[t]{@{}l@{}}User-involved design\\Psychological harm\\Acceptance and trust\\Emotional impact\end{tabular} & Psychological harm & \begin{tabular}[t]{@{}l@{}}Hierarchical decision-\\making\end{tabular} & \begin{tabular}[t]{@{}l@{}}Performance\\monitoring, fear\\of surveillance\end{tabular} &  \\ 
\hline
\begin{tabular}[t]{@{}l@{}}Veruggio et al. [53]\end{tabular} & Job loss & Privacy & Cybersecurity & & \begin{tabular}[t]{@{}l@{}}Psychological problems\\Emotional relationship\end{tabular} & \begin{tabular}[t]{@{}l@{}}Psychological problems\\Unpredictable\\behaviors\end{tabular} & Instruction conflicts &   & \begin{tabular}[t]{@{}l@{}}Rights and\\responsibilities\end{tabular} \\ 
\hline
\begin{tabular}[t]{@{}l@{}}Dubber et al. [31]\end{tabular} & Labor displacement  &  & & Transparency & Rights and well-being &   & Autonomy  &   & \begin{tabular}[t]{@{}l@{}}Accountability and\\responsibility\end{tabular} \\ 
\hline
\begin{tabular}[t]{@{}l@{}}Bartneck et al. [32]\end{tabular} &  & Privacy  & &   & \begin{tabular}[t]{@{}l@{}}Trust and fairness\\Psychological aspects\end{tabular}  & \begin{tabular}[t]{@{}l@{}}Risk in business\\Psychological aspects\end{tabular} &  &  & \begin{tabular}[t]{@{}l@{}}Responsibility\\and liability\end{tabular} \\ 
\hline
\begin{tabular}[t]{@{}l@{}}Berger and Rossi [56]\end{tabular} & Social impact  & \begin{tabular}[t]{@{}l@{}}Privacy and\\data governance\end{tabular}  & \begin{tabular}[t]{@{}l@{}}Privacy and\\data governance\end{tabular} & \begin{tabular}[t]{@{}l@{}}Explainability\\and trust\end{tabular} & \begin{tabular}[t]{@{}l@{}}Explainability\\and trust\end{tabular} &  &  & \begin{tabular}[t]{@{}l@{}}Profiling and\\manipulation\end{tabular}  & Accountability \\ 
\hline
\begin{tabular}[c]{@{}l@{}}Thiebes et al. [64]\end{tabular} &  &  &  & Explicability  &  & Non-maleficence & Autonomy &   &  \\ 
\hline
Liu et al. [65]    &  & Privacy   &   & Explainability  &   & \begin{tabular}[t]{@{}l@{}}Safety and\\robustness\end{tabular}  &  &  & \begin{tabular}[t]{@{}l@{}}Accountability\\and auditability\end{tabular} \\ 
\hline
\begin{tabular}[t]{@{}l@{}}Emaminejad and\\Akhavian [27]\end{tabular} &   & \begin{tabular}[t]{@{}l@{}}Privacy and\\security\end{tabular} & \begin{tabular}[t]{@{}l@{}}Privacy and\\security\end{tabular} & \begin{tabular}[t]{@{}l@{}}Explainability and\\interpretability\end{tabular} &  & \begin{tabular}[t]{@{}l@{}}Performance\\and robustness\\Reliability\\and safety\end{tabular} & &  & \\ 
\hline
\begin{tabular}[t]{@{}l@{}}European\\Commission [66,67]\end{tabular} & Societal well-being & \begin{tabular}[t]{@{}l@{}}Privacy and\\data governance\end{tabular} &  & Transparency  &  & \begin{tabular}[t]{@{}l@{}}Technical robustness\\and safety\end{tabular} & \begin{tabular}[t]{@{}l@{}}Human agency\\and oversight\end{tabular}  &  & Accountability \\ 
\hline
\textbf{Our research} & \textbf{Job loss}  & \textbf{Data privacy} & \textbf{Data security} & \textbf{Data transparency} & \textbf{Acceptance and trust} & \textbf{Reliability and safety} & \textbf{Decision-making conflict} & \textbf{Fear of surveillance} & \textbf{Liability} \\ 
\hline
\multicolumn{10}{l}{*sexuality, diversity, fairness, justice, beneficence, and nondiscrimination are not related to the AEC industry.}                                                                            
\end{tabular}%
}
\end{table}
\end{landscape}

\section{Systematic Review and Scientometric Analysis}
\label{sec:systemReview}

Numerous studies have been carried out to examine the ethical implications of AI and robotics in various domains.
However, little attention has been given to the ethical concerns arising from the development of these technologies in the AEC industry.
To address this gap, we conducted a systematic review to conceive the ethical dilemmas that emerge in AEC research. 
Several keywords and search criteria were first defined and used to collect relevant articles.
Then, the review process was performed following the PRISMA principles to extract the included articles and determine the ethical issues.
The scientometric analysis was also utilized to examine the keywords of the included articles.
Finally, the included articles were categorized based on their different research topics.

\subsection{Keywords and Flow Diagram}
\label{sec:keywords}

\begin{figure}[htbp]
    \centering
    \includegraphics[scale=0.54]{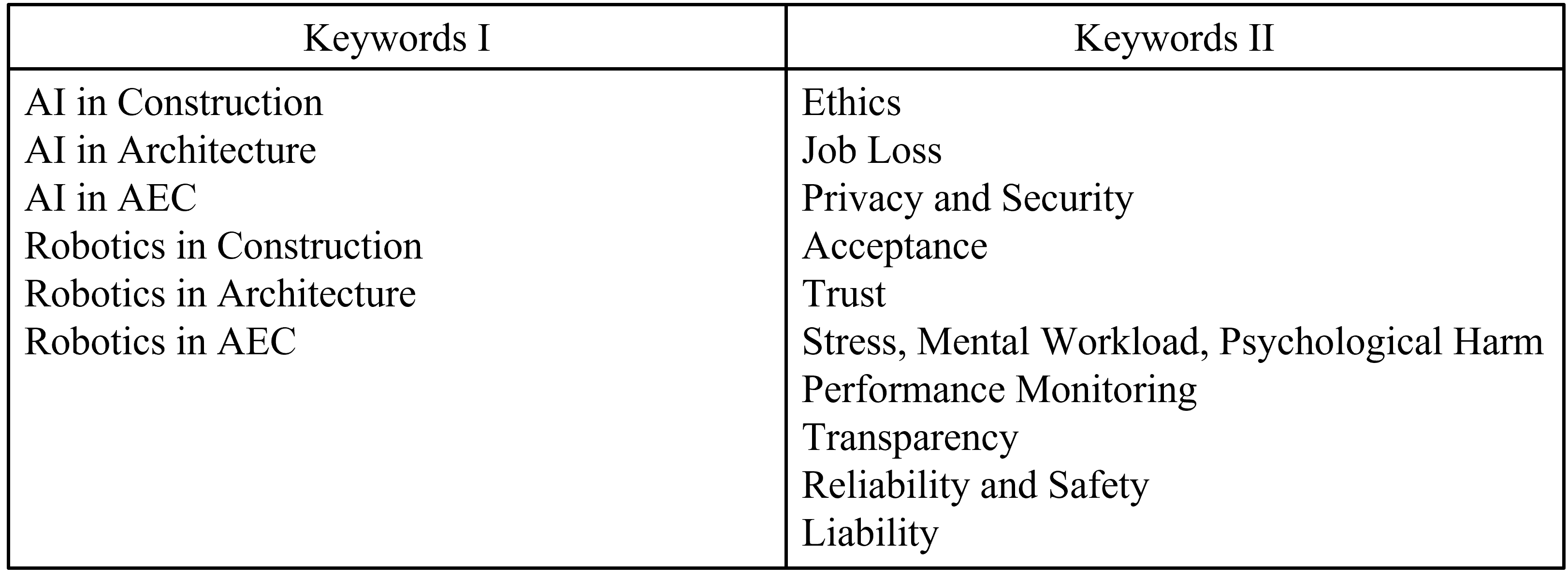}
    \caption{Keywords definitions with two groups}
    \label{fig:keywordsDef}
\end{figure}

To collect the relevant articles, we used the Scopus database to search for research publications.
Keywords were defined based on the ethical issues identified in Section \ref{sec:ethicsAIR} and were narrowed to the AEC discipline.
Covidence software was utilized to conduct the systematic review \citep{covidence_2022}.
Figure \ref{fig:keywordsDef} illustrates the definition of the keywords.
The search process was performed by choosing one phrase from Keywords I and one phrase from Keywords II in the title, abstract, and keywords fields.
For example, one of the search criteria was (TITLE-ABS-KEY("AI in construction") AND TITLE-ABS-KEY("ethics")).
In order to focus on recent research, we limited the results to publications from 2017 and only included journal articles, conference articles, and book chapters.
Furthermore, we also manually included articles from \emph{Automation in Construction}, \emph{Journal of Computing in Civil Engineering}, and \emph{Advanced Engineering Informatics} to expand the collection of AEC-related research.
Every set of search results was uploaded to Covidence for further analysis.
The search and manual scan process was conducted between 12/21/2021 and 1/12/2022.
Thus, this review did not cover the articles published after 1/12/2022.

\begin{figure}[htbp]
\centering
\includegraphics[scale=0.45]{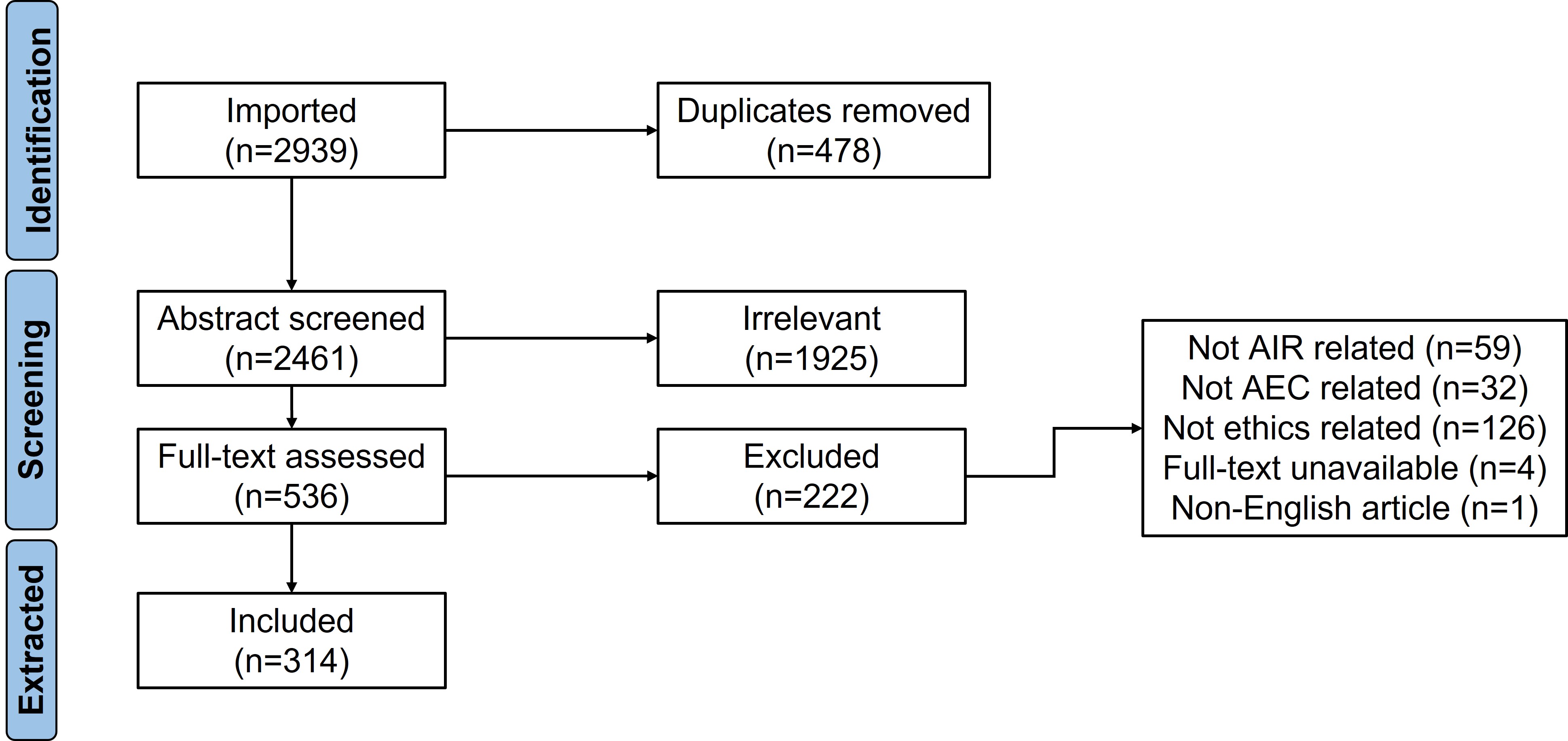}
\caption{Flow chart of the systematic review procedure}
\label{fig:Flow Diagram}
\end{figure}

Figure \ref{fig:Flow Diagram} shows the systematic review procedure and the results.
In the identification phase, we found 2,939 articles from the database and manual search process and imported them into Covidence software.
A total of 478 duplicated articles were removed automatically and resulting in 2,461 articles.
In the screening phase, the articles were first screened based on their title and abstract.
As a result, we included 536 relevant articles and excluded 1,925 irrelevant articles.
Next, the 536 relevant articles were assessed through a full-text review, and 314 of them were included while 222 of them were excluded.
Among the excluded articles, 59 articles were not related to AI and robotics, 32 articles were not related to AEC topics, 126 articles were not related to ethical issues, four articles were full-text unavailable, and one article was a non-English article.
Finally, in the extracted phase, all 314 included articles were analyzed based on ethical issues and further determined future research directions.

\subsection{Scientometric Analysis}
\label{sec:scientometric}


We conducted a basic scientometric analysis to explore the relationship between extracted articles.
It can be seen that research relating to ethical issues in the AEC industry had an upward trend in the last five years.
The number of articles with ethical factors increased from 18 in 2017 to 93 in 2021, which is approximately from 6\% to 30\% of the examined papers.
In January 2022, there have been 15 articles related to the topic published.
The number is expected to grow as new and additional publications have yet to be accounted for.



\begin{figure}[ht]
\centering
\includegraphics[scale=0.8]{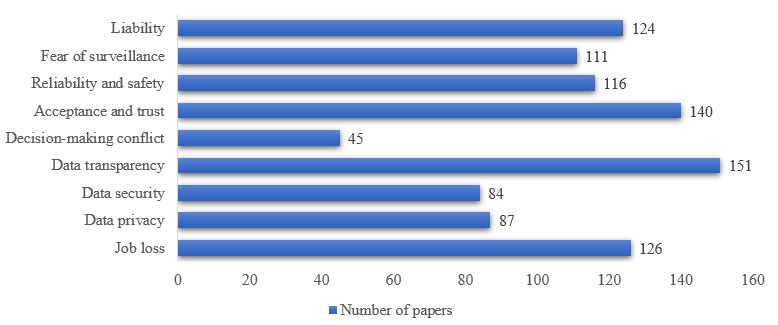}
\caption{Nine categories of ethical issues}
\label{fig:Categories}
\end{figure}

Figure~\ref{fig:Categories} shows the results of the full-text review.
According to the discussion in section \ref{sec:ethicsAIR} and Table~\ref{table:ethics}, we identify nine categories of ethical issues marked, including “job loss,” “data privacy," "data security,” “data transparency,” "decision-making conflict," “acceptance and trust,” “fear of surveillance,” “reliability and safety," and “liability.”
Among these, data transparency has the most related articles, 151 articles equal to the highest proportion of 15.35\%.
Acceptance and trust is the second relevance, with 140 articles.
Issues of job loss and liability have 126 and 124 articles, respectively.
Two other issues that have similar shares are reliability and safety and fear of surveillance, with 116 and 111 papers relatively.
The last three least concerned issues, according to our results, are data privacy, data security, and decision-making conflict.
Further explanation and concatenation for this distribution are discussed in subsequent sections.

\begin{figure}[htbp]
\centering
\includegraphics[scale=0.27]{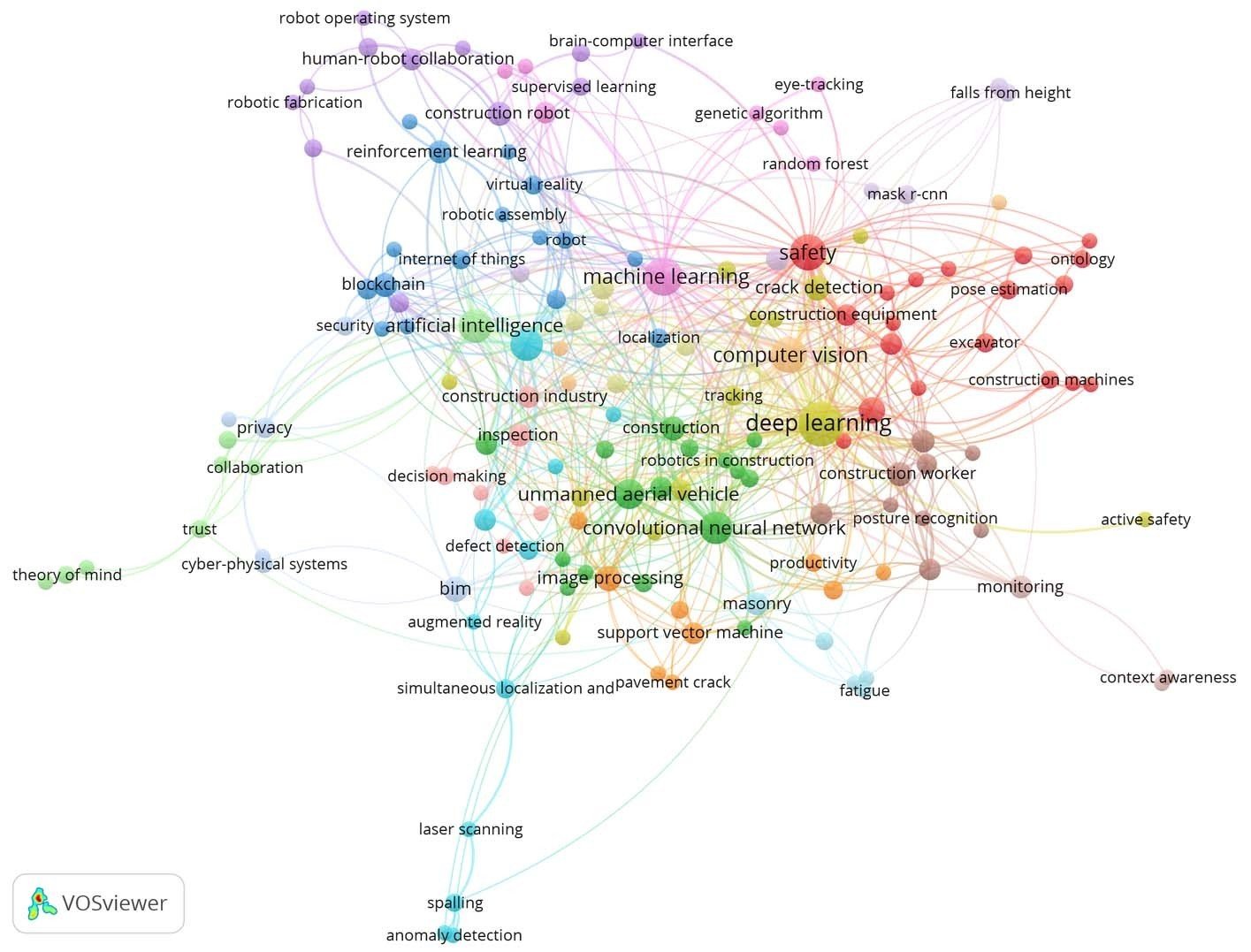}
\caption{Keywords co-occurrence based on bibliographic data}
\label{fig:KeywordsOccurrence_thres2}
\end{figure}

This meta-analysis is in compliance with the PRISMA guideline \citep{prisma_transparent_2022}.
The VOSviewer 1.6.18 was used to implement text mining and occurrence analysis.
In the output, each keyword is represented by a node, and there are various edges linking the nodes together to create the network of occurrence.
The edges have different transparency showing the strength of the connection between keywords.
Figure~\ref{fig:KeywordsOccurrence_thres2} shows a map of keyword occurrence based on bibliographic data (titles and abstracts) of collected articles.
A threshold of two occurrences was set for each keyword to create the network of keywords, and fractional counting was used.
Identical terms were merged, e.g., ML and machine learning, BIM and Building Information Modeling, and safety and construction safety.
As a result, among 1,150 keywords, 168 terms met the threshold and were selected to generate the map.
This map shows the main interests in all extracted papers.

The software categorized these words automatically into 16 clusters based on factors such as co-occurrence, similarity in definition, and usage.
AI-related terms were the most frequently used keywords, including “artificial intelligence,” “deep learning,” “computer vision,” “deep learning,” “machine learning,” and “convolutional neural network.”
Other frequently occurring terms were “building information modeling,” “construction,” “safety,” “ergonomics,” “facility management,” “construction worker,” and “monitoring.”
Our key term “ethics” was directly connected to “privacy,” “digitization,” and strongly related to “artificial intelligence.”
This network shows that ethical problems are not always discussed directly in AEC-related articles.
It is required to perceive ethical issues through other research topics as indicated by those most occurring keywords. 

\subsection{Research Topics}
\label{sec:researchTopics}
To further analyze the extracted articles, we categorized them based on the research topics.
We thoroughly reviewed each extracted article and determined the research topic.
Then, we grouped the articles with relevant topics and narrowed them down to 13 topics.
Table~\ref{table:topics} shows a list of 13 research topics along with the number of articles in each category.
These research topics are ergonomic risk analysis, physiological monitoring, localization and navigation, on-site monitoring, safety checking, scene reconstruction and progress monitoring, structure inspection and monitoring, assembly robot, robot and equipment control, cloud computing and security, design and planning, management and maintenance, and ethics.
For the ergonomic risk analysis category, the research focused on ergonomic safety \citep{zhao_applying_2021,chu_monocular_2020}, physical fatigue \citep{liu_effects_2021,zhang_jerk_2019}, or disability status prediction \citep{koc_integrating_2021}.
The data collection methods included vision system \citep{seo_automated_2021,zhang_ergonomic_2018}, wearable sensors \citep{nath_automated_2018}, or fusing vision and wearable sensors \citep{alwasel_identifying_2017}.
For the physiological monitoring category, researchers used sensors such as electroencephalography (EEG) \citep{aryal_monitoring_2017}, functional near-infrared spectroscopy (fNIRS) \citep{zhou_hazard_2021}, electrocardiography (ECG) \citep{hashiguchi_real-time_2020}, or eye-tracking \citep{li_identification_2020} to collect physiological data.
Such data was further used to monitor stress \citep{jebelli_eeg-based_2018}, mental fatigue \citep{li_identification_2020}, inattentiveness \citep{kim_predicting_2021}, the decision-making process of the human worker \citep{hu_application_2019}, or brain-robot interaction \citep{liu_brain-computer_2021,liu_brainwave-driven_2021}.

In the localization and navigation category, the research topics were image-to-BIM \citep{boroujeni_perspective-based_2017,wei_vision_2019} or sensors-to-BIM localization \citep{park_self-corrective_2017}, or robot localization and navigation \citep{mantha_robotic_2018,mantha_investigating_2022}, including mobile robots \citep{asadi_vision-based_2018}, drones \citep{nahangi_automated_2018}, and combination of mobile robots and drones \citep{kim_uav-assisted_2019,asadi_integrated_2020}.
One of the major methods developed in this category is the Simultaneous Localization and Mapping (SLAM) algorithm \citep{xu_occupancy_2019}, especially for indoor GPS-denied environments.
In the on-site monitoring category, the majority of research focused on tracking objects on construction sites, such as equipment \citep{kim_adaptive_2017,liang_vision-based_2019}, human \citep{konstantinou_adaptive_2019,wei_recognizing_2019}, human and equipment \citep{fang_automated_2018,lin_temporal_2021}, and materials tracking \citep{zheng_virtual_2020,zhou_image-based_2021}.
Computer vision \citep{liang_stacked_2018,luo_full_2020,xiao_development_2021} or sensors-based \citep{ma_li_smart_2017,rashid_activity_2020} were two main data sources for tracking and monitoring construction sites.
The applications of the on-site monitoring research were struck-by hazards detection \citep{son_real-time_2019,kim_proximity_2020}, proximity detection \citep{park_improving_2017,kim_remote_2019}, activity recognition \citep{luo_convolutional_2018,roberts_end--end_2019,langroodi_activity_2021}, productivity analysis \citep{kim_vision-based_2019,chen_automated_2020,kassem_measuring_2021}, trajectory or behavior prediction \citep{dong_proactive_2018,arslan_semantic_2019,cai_context-augmented_2020,tang_video-based_2020}, decision support \citep{zhou_dynamic_2019,garcia_decision_2021}, and monitoring system \citep{you_5g-based_2021}.
In the safety checking category, personal protective equipment (PPE) was detected by using vision-based \citep{fang_falls_2018,fang_detecting_2018} or sensor-based \citep{cho_machine_2018} methods to ensure worker safety.
Other research focused on risk and hazard analysis \citep{bigham_artificial_2018,tang_machine_2021}, safety rule checking \citep{kim_uas-based_2021,xiong_onsite_2019}, and accident analysis \citep{baker_ai-based_2020,cheng_text_2020} using AI algorithms.

In the scene reconstruction and progress monitoring category, research was mainly focused on point cloud and 3D model reconstruction \citep{shang_real-time_2018,kim_3d_2021}, component segmentation \citep{czerniawski_automated_2020,yan_automated_2021}, 4D BIM \citep{kropp_interior_2018}, digital twins \citep{akanmu_towards_2021}, progress monitoring \citep{lei_cnn-based_2019,pour_rahimian_-demand_2020}, and model updating \citep{wu_bayesian_2020}.
Reality capture and robotics technologies were applied to monitor the construction progress \citep{adan_autonomous_2020,kim_deep_2022}, which was further combined with blockchain technology to process payment automatically \citep{hamledari_construction_2021}.
The structure inspection and monitoring category consists of the most amount of the included articles.
The topics of this category included crack detection or segmentation for different types of structures, such as concrete surface \citep{alipour_robust_2019,chow_anomaly_2020}, steel frame \citep{kim_investigation_2021,perry_automated_2022}, masonry \citep{wang_automatic_2019}, ceiling \citep{wang_damaged_2020}, façade \citep{chen_geo-registering_2021}, bridge \citep{li_recognition_2017}, tunnel \citep{menendez_tunnel_2018}, and pavement \citep{zhang_deep_2018,shim_road_2021}.
Crack or defect classification for wood \citep{kamal_wood_2017}, sewer \citep{kumar_automated_2018,meijer_defect_2019}, or masonry \citep{dais_automatic_2021}, and crack analysis such as volume \citep{beckman_deep_2019}, seismic assessment \citep{xiong_automated_2020}, or spatial mapping \citep{ali_real-time_2021} were other research topics.
Different technologies, including mobile robots \citep{sutter_semi-autonomous_2018,mclaughlin_automated_2020}, drones \citep{zhong_assessment_2018,lin_bridge_2021}, cable robots \citep{hou_inspection_2020}, laser \citep{guldur_erkal_laser-based_2017,turkan_adaptive_2018}, computer vision \citep{liu_computer_2019}, or contact sensors \citep{kocer_inspection-while-flying_2019,gonzalez-desantos_uav_2020} were applied for inspecting and monitoring.
Finally, additional studies investigated bolt-looseness \citep{huynh_vision-based_2021}, steel bar quality inspection \citep{kardovskyi_artificial_2021}, and foundation pit or dam monitoring \citep{wu_rapid_2021,zhao_structural_2021}.

The next two categories are related to construction robotics.
In the assembly robot category, different types of construction robots were developed and deployed on-site to assist with assembly tasks.
These included masonry and bricklaying robots \citep{goessens_feasibility_2018,bruckmann_simulation_2021}, tile installation robots \citep{liang_teaching_2020,liang_trajectory-based_2022}, fabrication and timber robots \citep{kasperzyk_automated_2017,hack_structural_2020,wagner_flexible_2020,apolinarska_robotic_2021}, and 3D printing robots \citep{zhang_large-scale_2018,kontovourkis_robotic_2020}.
Moreover, repairing robots were another type of robot in the field to overcome repairing and assembly tasks at dangerous places, such as anchoring \citep{melenbrink_autonomous_2020}, rebar binding \citep{jin_robotic_2021,momeni_automated_2022}, and cable repairing robots \citep{xu_developing_2021}.
Lastly, the safety aspect between humans and robots on construction sites was also investigated \citep{you_enhancing_2018}.
In the robot and equipment control category, path planning methods for different construction tasks were studied, such as inspection \citep{phung_enhanced_2017}, joint filling \citep{lundeen_autonomous_2019-1}, masonry \citep{ding_bim-based_2020}, maintenance \citep{krishna_lakshmanan_complete_2020}, and crane \citep{hu_practicality_2021}.
Furthermore, remote control \citep{zhu_neurobehavioral_2021,koh_teleoperated_2021} or control methods \citep{groll_autonomous_2019,dadhich_field_2019,guzman_design_2019} were applied in the construction equipment, robot, smart infrastructure \citep{radmard_rahmani_framework_2019}, and hoist system \citep{lee_autonomous_2021} in the fields.
The human-robot collaborative control and digital twins were investigated to tackle controlling issues of the robots \citep{wang_interactive_2021,liang_bi-directional_2020}.

In the category of cloud computing and security, researchers have focused on federated cloud or federated learning applications \citep{li_federated_2021,erri_pradeep_blockchain-aided_2021,petri_coordinating_2017}.
Emerging technologies such as blockchain have been examined in the AEC industry \citep{ye_cup--water_2018,lu_exploring_2021}.
Finally, researchers also looked into the cybersecurity issues in construction, building, and smart cities \citep{turk_systemic_2022,woo_overview_2021,karaturk_security_2020}.
In the ethics category, the relevant ethical issues in AI and robotics were identified and discussed.
For instance, \citet{hatoum_developing_2020} studied the challenges of adopting robots in construction and how workers might be impacted.
\citet{schia_introduction_2019} identified the factors of human-AI collaboration in the construction industry.
Trust was found to be a key factor affecting AI and robotics in the AEC industry \citep{emaminejad_trustworthy_2022}.

In the design and planning category, several studies have focused on building design based on human behaviors or building performances \citep{van_ameijde_generative_2018,chokwitthaya_combining_2019}.
They have also developed layout planning methods for floor, window, facade, and structure layout, or automatic scheduling using different machine learning algorithms \citep{mangal_automated_2018,karan_intelligent_2019,agirbas_facade_2019,upasani_automated_2020,xu_optimal_2021,soman_automating_2022}.
In addition, price prediction, productivity analysis, clash resolving, and profit estimation were also investigated \citep{cao_prediction_2018,garcia_de_soto_productivity_2018,hu_clash_2019,bilal_guidelines_2020}, as well as the prediction of structural capacity \citep{ngo_integration_2021}.
In the management and maintenance category, researchers have investigated code checking and BIM element classification \citep{zhang_integrating_2017,koo_using_2019} and analyzed crew performance using fuzzy-based modeling \citep{raoufi_fuzzy_2018}.
The other research themes in the maintenance include maintenance classification, prediction, and staff assignment \citep{mcarthur_machine_2018,allah_bukhsh_maintenance_2020,mo_automated_2020}.
Lastly, the AI-based management research was also concentrated on infrastructure, facility, and project management \citep{soga_whole_2018,xu_cognitive_2019,das_blockchain-based_2022}.

\begin{table}
\caption {List of topics}
\label{table:topics}
    \centering
    \begin{adjustbox}{width=1.1\textwidth,center=\textwidth}
    \begin{tabular}{lll}
    \hline
       \textbf{Topic} & \vtop{\hbox{\strut \textbf{Number}}\hbox{\strut \textbf{of article}}}& \textbf{Notes} \\
       \hline
       Ergonomic Risk Analysis & 19 & Vision,   wearable sensors, or fusion\\
       & & Ergonomic safety, physical fatigue, or disability status prediction \\
       Physiological Monitoring &    9 &    EEG,   fNIRS, ECG, or eye-tracking\\   & &  Stress, mental fatigue, inattentiveness, or decision-making\\ & & Brain-robot interaction \\
       Localization and Navigation & 18 & Image-to-BIM, or Sensors-to-BIM localization\\& & Robot localization and navigation \\
       On-site Monitoring &    51 &    Vision   or sensors\\ & & Equipment, human, human and equipment, or object tracking\\     & & Activity and productivity analysis\\     & & Trajectory or behavior prediction\\     & & Decision-support and monitoring \\
       Safety Checking &    28 &    Vision or sensors-based PPE detection\\ & & Risk and hazard analysis, safety rule checking, or accident analysis \\
       Scene Reconstruction and Progress Monitoring &    19 &    Point cloud and 3D model reconstruction, or component segmentation\\ & & 4D BIM, digital twins, progress monitoring, or model updating\\   & & Automatic payment based on the progress \\
       Structure Inspection and Monitoring &    56 &    Mobile robot, drone, or cable robot\\ & & Laser, vision, or contact-based\\ & & \vtop{\hbox{\strut Crack detection or segmentation for concrete, steel frame, masonry,}\hbox{\strut ceiling, façade, bridge, tunnel, or pavement}}    \\ & & Crack or defect classification for wood, sewer, or masonry\\ & & Crack analysis such as volume, seismic assessment, or spatial mapping\\ & & Bolt-looseness, steel bar quality inspection, foundation pit or dam   monitoring \\
       Assembly Robot &    20 &    Fabrication   or 3D printing robot\\& & Masonry or tile installation robot\\& & Anchoring, rebar binding, or cable repair robot\\& & Human-robot collaboration safety \\
       Robot and Equipment Control &    25 &    Path planning such as inspection, joint filling, masonry, maintenance, or   crane\\ & & Remote control or autonomous excavator, bulldozer, or robot\\& & Human-robot collaborating control and digital twin\\   & & Smart structure or hoist system control \\
       Cloud Computing and Security &    17 &    Federated cloud or learning\\& & Cybersecurity in construction, building, infrastructure, or smart city \\
       Design and Planning &    24 &    Design   based on human behavior or building performance\\& & Floor, window, façade, or structure layout planning or automatic scheduling\\& & Price prediction, productivity analysis, clash resolving, or profit estimation\\& & Structural capacity prediction\\
       Management and Maintenance &    13 &    Code checking, BIM element classification\\& & Crew performance\\& & Maintenance classification, prediction, or staff assignment\\& &Infrastructure, facility, or project management \\
       Ethics &    14 &    Ethical   issues \\
       \hline
    \end{tabular}
    \end{adjustbox}
\end{table}

\section{Ethical Issues of AI and Robotics in AEC Industry}
\label{sec:issues}
We will discuss the nine ethical issues of AI and robotics and their implications in the AEC discipline.

\subsection{Job Loss}
\label{sec:job}

The introduction of AI and robots in various industries raises concerns about potential job loss, including AEC.
Workers are in fear of being replaced by AI or robots.
The AEC industry is resistant to emerging technologies because it is a well-established sector \citep{hatoum_developing_2020}, but still requires further improvement in terms of safety, productivity, and quality.
In this review, 126 out of 314 included articles are related to the job loss category.
Robots are used on construction sites to navigate around to collect data, register to the BIM model, monitor the construction progress, and perform tasks that were used to complete by human workers manually \citep{shang_real-time_2018,asadi_vision-based_2018,nahangi_automated_2018,park_framework_2019,kim_uav-assisted_2019,adan_autonomous_2020,asadi_integrated_2020,pour_rahimian_-demand_2020,hamledari_construction_2021,martinez_vision-based_2021,kim_deep_2022}.
Inspection robots have the ability to navigate environments that are hazardous or inaccessible to human workers, such as underfloor \citep{cebollada_mapping_2018}, bridge \citep{peel_localisation_2018,sutter_semi-autonomous_2018,morgenthal_framework_2019,lin_bridge_2021}, structure \citep{kocer_inspection-while-flying_2019,gonzalez-desantos_uav_2020}, cable \citep{hou_inspection_2020}, seismic damaged building \citep{xiong_automated_2020}, or facades \citep{chen_geo-registering_2021}.
Assembly robots and robot control systems are two types of robots that directly work on construction tasks \citep{melenbrink_-site_2020,liang_humanrobot_2021}.
On one hand, they can be used for 3D printing \citep{zhang_large-scale_2018,kontovourkis_robotic_2020}, timber \citep{wagner_flexible_2020,apolinarska_robotic_2021,kunic_design_2021}, masonry \citep{goessens_feasibility_2018,bruckmann_simulation_2021}, ceiling tile \citep{liang_teaching_2020,liang_trajectory-based_2022}, or rebar binding \citep{jin_robotic_2021}.
On the other hand, they can work autonomously on specific construction tasks, such as joint filling, drywall assembling, trenching, dozing, or excavating \citep{lundeen_autonomous_2019-1,groll_autonomous_2019,kim_task_2020,wang_interactive_2021}.
The increasing utilization of construction robots could lead to the elimination of human jobs.

Furthermore, AI and machine learning technologies are being applied to expedite or automate manual work.
For example, measuring the excavator productivity \citep{kassem_measuring_2021}, monitoring unmanned bulldozers \citep{you_5g-based_2021}, applying e-construction technologies to the highway project \citep{patel_evaluating_2019}, optimizing structure frames or building design \citep{mangal_automated_2018,karan_intelligent_2019,baghdadi_design_2020}, and managing construction project \citep{makaula_impact_2021}.
Inspection and monitoring of structures or infrastructure also reduce the manual inspection duty, such as bridges \citep{li_recognition_2017,jeong_sensor_2019}, pavement cracks \citep{hadjidemetriou_automated_2018,hoang_image_2018,zhang_deep_2018}, tunnels \citep{menendez_tunnel_2018}, concrete cracks \citep{turkan_adaptive_2018,zhong_assessment_2018,alipour_robust_2019,dung_autonomous_2019}, steel frame inspection \citep{kardovskyi_artificial_2021}, or safety regulation \citep{bigham_artificial_2018}.
As a result, the use of new AI and robotics technologies in AEC also shifts human workers' job duties to incorporate new processes.
Even if the number of jobs is not changing significantly, sometimes even creating new jobs, the existing workers still need to obtain new skills to fulfill the new job requirement.
This job displacement or reformation needs to be addressed to alleviate concerns \citep{tamers_envisioning_2020}.

According to a survey by \citet{kim_delegation_2022}, AEC professional practitioners have different levels of concern regarding robot adoption.
While some of them believe that robots will take over their job even if they were informed that their job would be the same, others think that robots will not outperform human workers due to the uniqueness and difficulty of high-skilled duties.
Even though the AEC industry is facing a skilled worker shortage and an aging workforce, the adoption of AI and robotics is still questioned by practitioners and labor unions.
Despite the effort of the job market reformation or redeployment \citep{van_wynsberghe_ethical_2022}, AI and robotics adoption in the AEC industry has critical challenges regarding job loss.

\subsection{Data Privacy}
\label{sec:privacy}

Emerging technologies have revolutionized management systems in the AEC industry. These technologies include Big Data, AI, Cloud Computing,  Internet of Things (IoT), Virtual Reality (VR), Augmented Reality (AR), Mixed Reality (MR), and Extended Reality (XR), as well as Robotic Process Automation (RPA) \citep{merlo_dynamic_2021, zhang_implementation_2022, schiavi_bim_2022, kido_assessing_2021, keung_data-driven_2021}.
The advancement enables intelligent supply chains and increases the flexibility of production lines and services, thereby improving customer satisfaction, especially when the speed of the internet continues to advance. 
Nonetheless, the issue of data privacy persists as a pivotal concern due to the prevalent digitization of data, which is subsequently stored on servers distributed across diverse geographical points. 
Consequently, internet-connected devices are susceptible to vulnerabilities that may occasionally entail profound implications for individuals.

Amidst the comprehensive assessment of articles undertaken within this study, over one-third related to data collection privacy issues.
Data collection is an essential step in construction automation, including object detection, tracking, and other technical measures \citep{kim_adaptive_2017,kong_quantifying_2018,mneymneh_vision-based_2019,chu_monocular_2020,tang_machine_2021}.
A typical process for AI-based progress monitoring in smart construction is to collect information, for example, visual data and sensor signals, at the beginning of work.
In order to collect such data, different tools and equipment are utilized, such as scanners, wearable sensors, and robot-mounted cameras.
These data can then be used as input for machine learning algorithms for various purposes in AEC research and practical work, particularly construction monitoring.

Data are usually collected for specific management purposes, including documentation of work monitoring and control \citep{nasiruddin_khilji_distress_2021, lei_cnn-based_2019}, safety \citep{mneymneh_vision-based_2019, yang_inferring_2019, cai_robust_2020}, risk analysis \citep{tang_machine_2021, nath_automated_2018}, and education and training \citep{fang_automated_2018, cao_prediction_2018, liang_trajectory-based_2022}.
These collected data may contain scenes irrelevant to the objectives and pose potential data privacy concerns. 
For example, workers might be captured without consent and prior notice, and data might include confidential project information accidentally.
This can result in inadequacy in protecting both individuals and project confidential information.
Stakeholders involved in a project may also have concerns about the unauthorized dissemination of project information.
On the other hand, potential data breaches may happen, particularly if the objects or the projects are associated with sensitive scenarios.
Those responsible for such illicit activities may exploit the stolen data for nefarious purposes.
Although not explicitly addressed in all of the articles reviewed, the issue of distinguishing and managing irrelevant data collected during construction projects remains unresolved.

\subsection{Data Security}
\label{sec:security}

The recent development of the internet and 5G technology enables high-speed data transmission, facilitating effective stakeholder communication. 
These technologies are instrumental in establishing networks, real-time data accessibility and transfer, and engagement with cyber-physical systems \citep{gupta_fusion_2021}.
Such endeavors turn the AEC industry into a complex production system \citep{erri_pradeep_blockchain-aided_2021}, which requires computers to work with extensive data.
While many AEC organizations still use conventional computer systems, traditional data storage methods such as local or company-wise servers are still popular.
However, this conventional storage habit generally poses problems for data security. 

Much research has raised concerns regarding data corruption during transferring and cyber security issues, such as snooping, information theft, viruses and worms, and hacking \citep{olatunji-2011, erri_pradeep_blockchain-aided_2021}.
Local devices on-site or in the home office may not have enough protection software against external offenses. There are many cases in which multiple users have credentials to access the same device or even share the credentials.
This usually resulted from a lack of computer security training or only conduct as a formality.
These reasons make computers vulnerable to data leakage, unexpected changes, or damage threats. On the other hand, the transferring process may be exposed to potential data leakage.
Additional data security issues are the integration of technologies, noises that decrease data quality, power outages, and untraceable data \citep{lu_exploring_2021,das_blockchain-based_2022}. 
Also, compliance with legal and contractual commitments does not protect security and safety breaches from data storage service providers, thereby leading to compromised data integrity and confidentiality \citep{ erri_pradeep_blockchain-aided_2021}.

Data security issues are also concerning in the AEC industry. First, engineers regularly need to synchronize the data from various software and different people when using multiple technologies in a construction project.
This collection and synchronization require multiple user access to computer storage and could lead to private data leakage \citep{lu_federated_2020, meijer_defect_2019}.
Data sharing can also be challenging when multiple control commands come from different access points without trusted parties.
Second, data quality can be insecure, untraceable, or damaged when having power outages during operation.
\citet{lu_exploring_2021} shows examples of BIM models in the shared cloud that can be tampered with and leave the data untraceable.
The study also stated that sensors could suddenly run out of power, and data quality could be corrupted because of unexpected noises during transferring. 

Blockchain technology is a potential solution to these problems.
It can enhance the transparency, traceability, and immutability of transferring data in the construction process \citep{lu_exploring_2021}.
Blockchains are defined as “tamper-evident and tamper-resistant digital ledgers” that are distributedly stored without central authority \citep{yaga2019blockchain}.
Zheng et al. described the blockchain’s architecture as blocks in sequence, and they contain an entire list of transaction records.
This technology can ensure some key characteristics of data, such as decentralization, persistence, anonymity, and auditability \citep{zheng2018blockchain}.
In the AEC industry, blockchain can enable consistency and hacking resistance in BIM designs \citep{das_blockchain-based_2022}.
\citet{xiong_blockchain-based_2022} developed a blockchain-based communication system to ensure the security and reliability of detection devices.
They confirmed that records saved in blockchain are auditable and traceable, as well as fairly shared among trusted devices.
Another application of blockchain is to secure automated construction payments in smart contracts \citep{hamledari_construction_2021}.
This method seems promising for payment with accuracy, efficiency, and time-saving.
It can be seen that the answer to the data security problem is within our grasp.
Thus, collaboration and coordination are left to trust among stakeholders and their teams.

\subsection{Data Transparency}
\label{sec:transparency}

Data transparency, explicability, explainability, and interpretability have been discussed in AI and robotics applications as one important aspect to improve human trust \citep{floridi_ai4people_2018,diakopoulos_transparency_2020,hacker_explainable_2020,bartneck_trust_2021,hacker_varieties_2022,emaminejad_trustworthy_2022}.
However, AI and robotics algorithms are usually opaque to end-users, i.e., a black box that generates outcomes \citep{berger_AI_2023,berger_addressing_2022,susser_invisible_2019}.
In particular, the advancement of deep learning makes AI systems more difficult to understand.
\citet{bartneck_trust_2021} argued that transparency differs from applicability, explainability, and interpretability.
For example, an AI algorithm can be open-sourced, making the entire code visible to anyone.
However, this information is not very useful for end-users since not everyone has adequate domain knowledge to understand the functionality and procedure of the code.
It is still nontransparent and prevents them from understanding how AI and robots produce results.

Explainable AI (XAI) \citep{adadi_peeking_2018} is a method that discloses information about the AI procedure and lets end-users easily understand how the AI came up with the result and what data it used.
Specifically, it can help users justify results, address incorrect results, and improve the system.
A cognitive architecture for collaborative robots can be helpful in explaining the robot's behavior \citep{cantucci_towards_2020}.
The General Data Protection Regulation (GDPR) developed by the European Union has regulations for AI to disclose personal data usage.
However, whether to enforce the AI system to explain how they made a certain decision is still under discussion \citep{hacker_varieties_2022}.
AI-generated explanations might not be suitable for increasing transparency since AI systems have to convince users to believe their explanations.
Even if AI systems generate a detailed explanation of how they reach that decision, the users could be concerned about if anything is hidden behind and not shared with them.
Therefore, it is better to have users reach a conclusion by themselves based on the information provided by AI \citep{diakopoulos_transparency_2020}.

In the AEC industry, AI and robotics transparency development is more challenging since most applications are developed by third-party companies, and end-users, i.e., construction practitioners, are not usually involved in product development.
The end-users need to be at the center of the AI system design in order to increase transparency for all stakeholders in the construction industry \citep{weber-lewerenz_corporate_2021}.
By introducing the ability of the AI system, transparency and trust of human-AI collaboration in construction can be improved \citep{schia_introduction_2019}.
In addition, the use of cloud computing, BIM, IoT, and blockchain technology has the potential to introduce more transparency in the AEC project since all stakeholders will have the ability to monitor the progress in real-time \citep{ye_cup--water_2018,bello_cloud_2021,woo_overview_2021}.

The application of XAI in the AEC discipline is still new, and not been many research efforts on this topic yet.
\citet{emaminejad_trustworthy_2022} studied trustworthy AI and robotics research and identified XAI as an essential concept for the AEC industry.
Similarly, \citet{love_explainable_review_2022} reviewed the XAI literature and suggested future research directions on XAI in construction, including stakeholder desiderata and information fusion.
An evaluation framework was proposed for construction stakeholders to evaluate the outcome of XAI adoption \citep{love_explainable_framework_2022}.
Interpretable machine learning, i.e., model interpreter, was developed in the applied machine learning guidelines for adopting the AI system in construction projects \citep{bilal_guidelines_2020}.
Even with a black-box AI system, using an interpretability analysis approach to evaluate trustworthiness can still confirm transparency in AEC applications \citep{liang_integrating_2023}.

\subsection{Decision-making Conflict}
\label{sec:decisionConflict}

Decision-making conflicts occur when AI and robots make decisions differently from humans in the same situation.
In decision-making scenarios, human workers rely on their experiences and the current situation, whereas AI systems make decisions based on the trained model and the current situation.
As a result, they might reach contradictory decisions despite having the same data and conditions.
In such cases, resolving these conflicts and making the final decision is more challenging for people because of the conflicted recommendations, and their decision may be influenced by the AI system \citep{susser_invisible_2019}.
These conflicts between humans and AI systems also decrease the team trust dynamic \citep{schia_introduction_2019}.
In addition, most AI systems do not consider the emotional or relational aspects of collaborating workers during the decision-making process, which are important to the performance of human workers \citep{van_wynsberghe_ethical_2022}.
AI systems can also be biased in making decisions due to the training data, which emphasizes the need for fair AI systems \citep{fernandez_inclusive_2019}.

In the AEC industry, the decision-making process is more complex as it involves large budgets and multiple stakeholders \citep{hu_application_2019}.
For example, BIM clash resolution requires gathering all stakeholders to determine how to resolve clashes across different models.
It will result in additional costs if the conflict happens in the construction phase.
Using AI systems can help human workers or managers analyze big data and make decisions in AEC projects \citep{hooda_emerging_2021,woo_overview_2021,bilal_guidelines_2020}, e.g., BIM design clashes resolution \citep{hsu_knowledge-based_2020}.
Decision support systems are considered one of these AI systems, which use environmental data to predict the outcome and make recommendations, such as tunnel boring machine operation \citep{garcia_decision_2021}, fatal accidents prediction \citep{choi_machine_2020}, equipment residual value prediction \citep{shehadeh_machine_2021}, and schedule look-ahead method \citep{soman_automating_2022}.
Data collection can be achieved by deploying mobile robots in built environments and later used in the decision-making process \citep{mantha_robotic_2018}.
Furthermore, path planning methods for robots or heavy equipment provide recommendations for human operators or collaborators to determine the route \citep{yousefizadeh_trajectory_2019,kayhani_heavy_2021,lee_autonomous_2021,wang_interactive_2021}.

In the design and planning category, AI systems are commonly involved in the process of assisting human workers in making decisions.
\citet{van_ameijde_generative_2018} utilized a generative design method in architecture and building design processes and determined design decisions based on site and context conditions.
\citet{karan_intelligent_2019} developed an intelligent designer to design building windows based on client needs and expectations.
AI systems can make design decisions and collect feedback from clients.
Predictions using AI during the design process support designers and managers in evaluating the design before making the final decision \citep{cao_prediction_2018,chokwitthaya_combining_2019,zhang_optimal_2019,chakraborty_novel_2020,pan_bim_2020}.
However, decision-making conflicts still happen and are difficult to resolve if AI systems come up with recommendations that are contrary to the usual decisions made by human workers.

Researchers have been working on robot decision-making modules that follow ethical guidelines, such as the cognitive architecture \citep{cantucci_towards_2020}.
This architecture not only assists the decision-making process but also increases transparency in human-robot collaboration.
Decisions made by robots can be easily altered from ethical to aggressive using an ethical layer before the robot controller \citep{vanderelst_dark_2018}.
On the other hand, involving human workers in the AI system is one way to resolve decision-making conflicts, known as human-in-the-loop decision-making \citep{diakopoulos_transparency_2020}.
Before arriving at a final decision, human workers can assess the suggestions made by the AI system at each step and provide comments to improve them.
Through this process, a mutually agreed decision can be reached while increasing the transparency of the AI system.

\subsection{Acceptance and Trust}
\label{sec:acceptance}

Trust in human-robot or human-AI collaboration plays a crucial role in the successful deployment, especially for team performance and safety.
When introducing emerging technologies to industries, it is necessary to ensure that human workers trust and accept them.
\citet{charalambous_trust_2022} pointed out that there was limited research on understanding trust development in industrial human-robot collaboration.
Previous studies have argued that human workers may be forced to accept new technologies \citep{van_wynsberghe_ethical_2022}, implying a negative relationship between humans and AI systems.
To establish trust, \citet{bartneck_trust_2021} proposed five principles for trustworthy AI, according to European ethical principles for AI, which are non-maleficence, beneficence, autonomy, justice, and explicability.
AI and robot systems should not hurt human workers.
Instead, AI and robot systems should benefit workers while preserving their rights and authority.
Fairness and explicability encourage human workers to understand the outcome of AI systems, which is related to data transparency and decision-making conflicts.
The objective of these principles is to build human workers' trust and acceptance when using or working with AI and robots in the workplace.

Measuring trust in human-robot/AI collaborative teams is one way to evaluate whether human workers accept these technologies as tools or collaborators.
Existing trust-related research in human-robot collaboration mainly utilized questionnaires at the end of the collaboration to determine the trust level.
For example, \citet{charalambous_development_2016} used a psychometric scale to evaluate human trust after human-robot collaboration in the industrial workplace.
The trust level changes are not considered during the process.
Recent studies have focused on modeling trust dynamics during human-robot collaboration \citep{jessie_yang_toward_2021}.
\citet{guo_modeling_2020} designed a personalized trust prediction model using Beta distribution to represent the worker's temporal trust in human-robot collaboration.
Additionally, physiological or psychophysiological signals, such as heart rate or EEG signals, can be used to measure trust with non-invasive devices \citep{ajenaghughrure_measuring_2020,hu_application_2019}.
\citet{hopko_trust_2022} tested the use of functional near-infrared spectroscopy (fNIRS) to link the brain neural response to trust in human-robot collaboration.
\citet{shayesteh_workers_2022} developed a wearable EEG-based system to continuously monitor the worker's trust level in construction human-robot collaboration.

Trustworthy AI and robotics have also been studied in the AEC discipline \citep{emaminejad_trustworthy_2022}.
The trust issue of applying machine learning algorithms to construction projects has been discussed \citep{bilal_guidelines_2020,schia_introduction_2019}.
When human workers lack a clear understanding of how AI works, they will not trust the AI system.
\citet{you_enhancing_2018} proposed a Robot Acceptance Safety Model (RASM) in order to measure the preserve safety in construction human-robot collaboration.
The introduction of cloud computing to the AEC industry has raised challenges of trust and psychological discomfort, such as concerns about data leakages \citep{bello_cloud_2021}.
Blockchain technology can be a solution to ensure data security as well as trustworthiness \citep{lu_exploring_2021}.
Furthermore, \citet{le_artificial_2022} designed an AI-based framework to ensure the privacy and fairness of IoT communications so that trust can be built.

\subsection{Reliability and Safety}
\label{sec:reliability}

Although there have been efforts to leverage AI and robotics in AEC in the last decades, their reliability in real-world applications is still an ongoing issue.
One of the biggest challenges to building trust in AI applications (e.g., computer vision) is due to the lack of common and objective criteria to validate the robustness of algorithms in the AEC field \citep{paneru_computer_2021,xu_computer_2021,fang_automated_2018,xuehui_dataset_2021}. 
For example, \citet{fang_computer_2020} confirmed that it is still challenging to benchmark the performance of prior AI works that have been trained and tested with different datasets and metrics in architecture, engineering, and construction. 
Such absence of common and objective criteria has resulted in a reliability issue associated with questioning the robustness of computer vision solutions in different real-world settings (e.g., various building types and different construction sites). 
The other pressing challenge towards trustworthy AI applications is the lack of explainability and interpretability \citep{fang_computer_2020-1,zhang_interpretable_2018}. 
They are often regarded as a blackbox, which means that the details of algorithms (e.g., feature engineering) are typically non-interpretable. 
This is basically relevant to the data transparency issue (section \ref{sec:transparency}), but it can further deteriorate as reliability issues of AI-based systems and their outcomes. 
Providing adequate explanations or justification is critical for the end users to be able to judge the trustworthiness of AI-based systems in architecture, engineering, and construction applications \citep{AKINOSHO_deep_2020}. 
In this regard, for example, a growing literature in the construction domain \citep{emaminejad_trustworthy_2022,naser_engineers_2021} identified the potential benefits of explainable AI to enhance user-friendly visual representation, such as LIME \citep{ribeiro_should_2016}, SHAP \citep{lundberg_unified_2017}, and Grad-CAM \citep{selvaraju_grad_2020}. 

Over the past decade, AI has greatly contributed to the field of robotics in architecture, engineering, and construction (e.g., autonomous construction robots). 
AI is essential for autonomous robotic controls since they need to be capable of making decisions and performing tasks like human experts maneuvering equipment taking account of various environmental factors. 
Although data collection robots (e.g., Unmanned Ground Vehicles (UGV), Unmanned Aerial Vehicle) equipped with AI have been employed for autonomous monitoring and inspection purposes, it is not trivial to deploy construction robots (e.g., excavators) that are fully capable of autonomously performing tasks and actions in the real-world settings. 
Recently, an AI-based autonomous excavator that can operate efficiently, robustly, and with generalizability has been proposed by \citet{zhang_autonomous_2021}. 
For their system, multiple sensors were employed (e.g., RTK, LiDAR, camera), while reinforcement learning and data-driven imitation learning were used for the planning module.
It was reported that their AI-based system performed 90 percent as efficiently as a human expert in digging and dumping tasks, but their system has only been tested indoors with predefined materials and repetitive tasks without nearby obstacles.
As such, the AI-based autonomous control of construction robots in prior works has primarily focused on tasks or actions in a controlled or simulated testbed \citep{jud_autonomous_2019,sandzimier_data_2020,sotiropoulos_model_2019,sotiropoulos_autonomous_2020}, and thus safety concerns and risks from AI-based robotics decision-making process have not been rigorously studied in the naturalistic settings.
It is noted that it is not trivial to leverage AI for autonomous robots performing various real construction tasks (e.g., digging, trenching, loading, dumping, leveling, compaction, rock removal, driving) in dynamic, complex, and suboptimal conditions (e.g., rain, excessive dust, extremely high and low temperature, ground slope, bumpy terrain, soil hardness, nearby obstacles, underground utility, and nearby robots and workers).
Facing such challenges is quite different from manufacturing robots that perform relatively simple and repetitive tasks in structured and controlled work environments, which raises safety concerns and risks from AI-driven robotic decision-making in the architecture and construction field.

\subsection{Fear of Surveillance}
\label{sec:surveillance}

The fear of surveillance represents a consequential ethical concern following the increasing deployment of cameras and sensors in the AEC industry. 
While camera systems have become essential on construction sites to collect visual data, the output quality of processing these visual data still heavily depends on the environment, such as light, weather, and the angle of observation \citep{son_integrated_2021, jin_robotic_2021}.
Besides visual sensing technologies, radio-frequency identification (RFID), global positioning system (GPS), radar, and laser scanners \citep{son_integrated_2021} are also being utilized to monitor construction activities. Both cameras and sensors have shown effectiveness in collecting task- and operation-level progress on the construction site. This includes tracking physical building elements and its quality \citep{mneymneh_vision-based_2019, chow_anomaly_2020, kropp_interior_2018}, worker and equipment activities and their interaction \citep{xiao_vision-based_2021-2,xiao_development_2021}. The worker activities and interaction can be analyzed to provide safety \citep{son_real-time_2019,luo_real-time_2020}, violation prevention \citep{guo2018real,zheng2019violation,yan2022deep}, and workplace behavior monitoring \citep{guo2018real,son_detection_2019}.

Despite the wide applications and benefits of sensor applications on-site, ethical concerns related to the constant surveillance of workers have been raised. The existence of the cameras and awareness of being monitored can influence workers' morale and work efficiency \citep{nishigaki_infographics_2019}.  
In contrast to mechanical processes, workers engage in tasks that embrace various motions and postures during their work activities. 
Awareness of continuous surveillance through cameras can cause a decline in employees' psychological and emotional well-being, engendering a sense of mistrust.
In other circumstances, open offices can foster positive attitudes in employees. Many new technology companies are examples of establishing a free and inspired workspace and significantly motivating their staff to develop creative ideas.
Practice shows that these companies have attained many achievements with their workstyle \citep{thoring2019inventory}.

On the other hand, as mentioned in section \ref{sec:security}, the risk of data being unsecured exists.
The more information is captured and stored, the higher the chances that information is being threatened.
The issues of data privacy and security have been discussed in previous sections.
As a result, it is understandable that people cannot achieve their best performance while fearing unprotected data privacy. Often, workers are not scared of constant surveillance, but they are afraid of compiled information being used against them \citep{shier_supervisor_2021}.

\subsection{Liability}
\label{sec:liability}

Machine learning and deep learning have driven digital transformation in construction by detecting and predicting progress, safety, and quality issues through visual and text data \citep{weber-lewerenz_corporate_2021}. Besides expanding the number of objects to detect, higher accuracy has been another main objective to pursue for practical reasons. Without high enough accuracies that managers could believe, the liability of these learning-based systems in practice could be doubtful. Liability refers to the responsibility of an individual or group that might be held accountable \citep{erri_pradeep_blockchain-aided_2021, tschider_regulating_2018}. Even though research shows that 94.3\% accuracy in detecting worker poses \citet{son_detection_2019}, 93.2\% accuracy in detecting equipment \citet{xiao_vision-based_2021-2}, and 93\% accuracy for detecting loosened bolt connection \citep{huynh_vision-based_2021}; the numbers might not be representative in practical settings without comprehensive onsite testing with practitioners to develop a liability system across the applications and users.

Liability issues could also happen in cloud-based systems such as BIM coordination platforms. As software resolves interoperability issues better, different contractors can access the federated model conveniently with feasible permission settings. However, cases show that bidding based on BIM tools could lead to undervaluing and cause liability issues without a contractual agreement on responsibility definition \citep{erri_pradeep_blockchain-aided_2021, tschider_regulating_2018}. In such cases, liability needs to be defined clearly through the BIM execution plan and permissions.

Another solution to enforce liability and security could be through blockchain applications \citep{parn2019cyber}, which shows the potential to limit the liability to only relevant stakeholders by storing data within the blockchain-based framework \citep{erri_pradeep_blockchain-aided_2021,yaga2019blockchain,zheng2018blockchain,xiong_blockchain-based_2022}. Blockchain has been integrated into other software, such as CAD, to maintain an immutable record of the modeling history \citep{dounas2018cad}. \citet{zheng2019bcbim} developed the bcBIM framework that can track the modifications in BIM data by integrating mobile cloud technology and blockchain. \citet{nguyen2019blockchain} also suggested that when blockchain is integrated with BIM in construction, project management may attain security, transferability, ownership, and liability.

\section{Discussion}
\label{sec:discussion}

The adoption of AI and robotics in the AEC industry has great potential in enhancing safety, efficiency, and productivity.
However, such advancement raises ethical challenges that require attention and solutions.
In this review, we have identified nine key ethical issues of AI and robotics in AEC: job loss, data privacy, data security, data transparency, decision-making conflict, acceptance and trust, reliability and safety, fear of surveillance, and liability.

Job displacement due to the automation of certain tasks is a significant concern.
As AI and robots take on repetitive and physically demanding tasks, there is a risk of job loss for human workers, especially for AEC workers.
Ethical considerations in this narrative should focus on providing support for reskilling and upskilling to ensure a smooth transition for affected workers, reducing the social and economic impact.

Data is an essential part of AI and robots.
In the AEC industry, data are collected and used to create various datasets for training and testing emerging systems.
Ensuring data privacy and security is important to protect sensitive information from unauthorized access and breaches.
Ethical AI implementation must include robust cybersecurity measures to maintain data integrity.
Furthermore, transparency in AI, such as explainable AI, is crucial for increasing workers' trust and for identifying and mitigating potential biases.

Since the AEC projects are large-scale and long-term, the decision-making process is highly complex and requires the involvement of multiple stakeholders.
Decision-making conflicts between AI and human workers may occur.
How to address these conflicts and establish a harmonious human-robot collaboration is the core of this ethical issue. 
On the other hand, acceptance and trust are vital for introducing AI and robots to AEC workplaces.
A poor trustworthy relationship between humans and AI will decrease the outcome of the collaboration.
Dispelling fears of surveillance can help build acceptance and trust in emerging systems.
In addition, the ethical development of AI and robotics in AEC must consider reliability and safety.
AI algorithms and robotics systems should be thoroughly tested and validated to minimize the risk of accidents and ensure the well-being of human workers.
Finally, since AI and robots have become more autonomous, concerns about liability and responsibility have emerged.
Clear guidelines and frameworks are necessary to determine accountability in the event of accidents or failures of AI and robot systems.

\section{Future Research Direction}
\label{sec:future}

Based on the discussion of ethical issues, we propose the following future research directions to encourage research efforts in this important field.




\subsection{Human-Robot Collaboration Safety}
The reliability and trustworthiness of AI-based robot and equipment controls in terms of safety are of the utmost importance for field deployment. 
In this context, future AI should be trained not only in automation skills but also in aspects related to safety regulations.
The advance of AI allows to reduce the dependence of the human operator on decision-making, but the dynamic and error-prone field environment is still challenging for relying on the AI system in the real world \citep{lee_challenges_2022}. 
Thus, how robustly involved human operator intervention in AI-based robot and equipment controls would be critical to improving the system reliability and safety.
Although higher levels of automation could be theoretically possible for decision-making, robust human intervention is still needed depending on in-situ conditions.
In this regard, human-centered research (e.g., coexistence, cooperation, collaboration) that can complement the AI-driven decision-making process while reducing safety concerns and risks needs to be conducted more by future researchers.

\subsection{Explainable AI}
AI algorithms used in construction should be transparent and explainable so stakeholders can understand how decisions are made. Transparency and explainability are particularly important for decisions regarding design, safety, and quality issues. Current deep learning frameworks or large language models often generate results from a black box where the reasoning relies on human interpretation. The interpretation can still be subjective, experience-based, or even purely speculative. For example, generative designs for buildings can create valid layouts without instructions, while the reasons behind the creation are more critical during evaluation. In construction safety, the violation detections from image captioning can be hard to understand without visualization of how the algorithms generate the result. Safety issues are related to worker's life, and quality issues could lead to rework. The decision-making process needs to be understandable and transparent. Developing explainable AI algorithms and standards for transparency in construction AI systems is crucial in the future.
Although studies have investigated the application of explainable AI methods in the construction domain's AI problems, the explainability of these methods might only satisfy some stakeholders' requirements. There is still a need to develop explainable AI that best suits construction problems at different stages and from stakeholders' perspectives. 

\subsection{Trust of AI and Robotics}
In the field of human-robot trust in AEC, the potential for future research to revolutionize the industry through innovative approaches is enormous. 
Future studies can explore designing AI and robots that exhibit trustworthiness on construction sites.
This includes enhancing their robustness and adaptability to dynamic environments and creating a foundation of unwavering reliance.
Another research direction can investigate designing AI and robots that can naturally generate trustworthiness in social interaction scenarios.
For example, improving the robots’ emotional intelligence and responsiveness and building a symbiotic human-robot relationship based on trust.

Focusing on the foundations of trust, future research can examine the profound implications of transparency.
Identifying which aspects of transparency can significantly shape trust and leveraging them in construction robot design can potentially lead to more reliable interaction paradigms.
In addition, the roles of safety and user control as pivotal trust drivers deserve to be explored.
Investigating effective methods to communicate safety information to humans and maintain user control in increasingly construction robots aligns seamlessly with the overarching goal of trust-building.
Moreover, it becomes critical to acknowledge the interplay of human factors in trust formation.
Individual differences in personality and cognitive style significantly influence trust perceptions.
Future studies may delve into robots that can adapt to these human factors, thereby fostering trust on a more personal level.

The ethical dimensions of trust in automation also require a full investigation.
As robots gain autonomy, the ethical implications of trust become even more prominent.
Understanding their ethical challenges, such as potential misuse or malicious intent, is crucial to deploying robots responsibly and carefully. 
These diverse research trajectories converge to form a rich research theme of future human-robot trust.
By connecting engineering developments with empirical insights, these research efforts have the potential to reshape the AEC industry and the fabric of human-robot interactions.
As AI and robots integrate into AEC workplaces, continued research remains critical to foster trust and prompt harmonious coexistence between humans, AI, and robots.

\subsection{Cybersecurity}
The AEC industry has been rapidly progressing towards the implementation of digital technologies with a particular emphasis on AI and robotics to enhance various processes across all the life cycle phases from inception till the end of life. Due to the inherent nature of this digitalization drive and innovative transformations, different stakeholders are directly and indirectly exposed to cybersecurity risks. In the context of the AEC industry, cybersecurity can be simply defined as means and methods to safeguard digitally connected resources, including physical infrastructure (e.g., automated equipment) and humans (e.g., workers on site) against potential wrongdoings (or cyberattacks). Although cybersecurity has been increasingly reported as a major concern due to these digital transformations in the AEC industry \citep{yoders_cybersecurity_2022}, there has been very limited research conducted within the community \citep{mantha_cybersecurity_2021}. Recent studies also emphasize the need to adopt and customize existing cybersecurity frameworks, tools, and methods in the context of the AEC industry, which has unique characteristics and challenges compared to other industries \citep{turk_systemic_2022,mantha_cyber_2021,mantha_assessment_2021,garcia_understanding_2022,Garcia_construction_2022}. A similar strategy was also warranted for other industries, such as manufacturing and critical infrastructure \citep{wu_cybersecurity_2018}. To address these gaps, future research can focus on AEC-specific investigation, exploration, adoption, and development of standards (e.g., data protection), policies (e.g., privacy protection), frameworks (e.g., zero trust security framework), tools (e.g., intrusion detection), methods (e.g., penetration testing), and mechanisms (e.g., role-based access control).

\subsection{Bias and Fairness in Decision-making}
Current AI systems are widely used in different construction applications, and these AI systems all require historical data to train. Since most of the research cannot claim that they have established a universal and comprehensive dataset, especially for construction applications, these systems are highly likely to learn and perpetuate biases in the data. These deep networks often make confident but incorrect predictions when tested with outlier data not seen in their training distributions. The violation report data could limit construction safety detection, which often focuses on falling from height. The same situation could happen in quality detection systems where the data distribution of the categories restricts the AI systems. Future AI systems could be improved by developing methods to mitigate biases during decision-making or automatically detect them and ensure the user could avoid mindlessly following potential discriminative decisions. For example, methods that compute likelihoods with deep generative models (DGMs) for outlier detection from unseen data. At the end of the day, the final decision maker needs to be well-informed of what might be missing from the AI systems.

\subsection{Responsibility Framework}
As AI systems become more autonomous, it will be essential to determine who is responsible for their actions. Traditional liability issues in construction have been largely covered in the contract to prevent litigation. While new technologies such as AI are being introduced to the process in many steps, the industry must introduce a standardized framework to regulate the effect of each decision-making. Several issues require further investigation to define a responsibility framework. For example, the responsibility of the decision-maker or the AI system. The decision-making process becomes not only related to the decision-maker but also to the AI system creation process. If the responsibility is related to the AI system, it could be developed by multiple people and involve multiple systems. The responsibility of the verification process of the AI system and the requirement of validating the decision through comparisons of different systems need to be defined. These topics require attention, and future work could focus on developing frameworks for assigning responsibility and accountability in AI systems used in construction.

\subsection{AI and Robotics Standards}
Current industrial standards of collaborative robot safety, ANSI/RIA R15.06 \citep{ansi_r15_2012} and ISO/TS 15066 \citep{iso_isots_2016} have identified four types of operations to control injury risks for collaborative robots to work with workers, including safety-rated monitored stop, hand guiding, speed and separation monitoring, and power and force limiting.
While these operations are effective in ensuring worker safety within designated workspaces, implementing them on construction sites poses unique challenges.
Unfortunately, the existing safety standards do not address the specific workspaces found on construction sites.
For example, demolition and masonry robots can pose injury hazards for construction workers.
Ineffective and careless operations of remote robotic devices on jobsites can potentially yield severe injuries.
To address these concerns, the Occupational Safety and Health Administration (OSHA) recommends that users review robot systems and applications in accordance with manufacturer requirements, industry standards, and relevant OSHA regulations \citep{osha_industrial_2021} for general industry and/or construction.
However, while the safety criteria of existing standards may apply to certain construction scenarios, they may not fully align with the specific requirements and concerns of construction workers. 

On the other hand, construction workers can also benefit from remote functions and supervise robots from a safe distance using dedicated remote controllers.
Collaborative robots on construction sites can serve as assisting tools to provide the required force, helping construction workers avoid potential injuries associated with certain tasks.
Given the increasing prevalence of robotic devices in construction applications involving extensive human-robot interaction, there is an urgent need for a comprehensive safety standard that specifically addresses the unique safety considerations of construction sites.
Such a standard would ensure the safe integration and operation of robots while protecting the well-being of construction workers. 

\section{Conclusion}
\label{sec:conclusion}
This research has investigated the issue of ethics concerning the adoption of AI and robotics in the AEC industry.
Through a systematic review of the literature, we have highlighted nine ethical issues requiring attention as these technologies continue to emerge in the AEC domain.
Seven future research directions are suggested to further research efforts in the field of AI and robotics in AEC with a specific emphasis on ethics.

This research has three limitations.
First, since the ethical topic is new to the AEC discipline, the keyword search method is difficult to collect relevant literature.
The data collection process heavily relies on manual collection (snowballing), which might not cover all relevant articles and cause bias. Nevertheless, the tremendous expertise of the authors and the relevant expert feedback loop considered largely tackle this concern.
Second, since the literature search data collection was conducted between December 2021 and January 2022, more recent studies in the AEC field are not covered in this systematic review.
Particularly, the emerging large language model (LLM) raises critical ethical concerns in every discipline.
Third, this research is limited to the literature published in English only, which has the potential to overlook relevant literature published in other languages.



\bibliographystyle{elsarticle-num-names} 
\bibliography{ref}





\end{document}